\documentclass{vgtc}                          

\ifpdf
  \pdfoutput=1\relax                   
  \usepackage{graphicx}                
  \DeclareGraphicsExtensions{.pdf,.png,.jpg,.jpeg} 
\else
  \usepackage{graphicx}                
  \DeclareGraphicsExtensions{.eps}     
\fi%

\graphicspath{{figures/}{pictures/}{images/}{./}} 
\usepackage{paralist}
\usepackage{microtype}                 
\PassOptionsToPackage{warn}{textcomp}  
\usepackage{textcomp}                  
\usepackage{mathptmx}                  
\usepackage{times}                     
\usepackage{cite}                      
\usepackage{tabu}                      
\usepackage{booktabs}                  
\usepackage{balance}
\usepackage{float}
\onlineid{1072}

\vgtccategory{Application or Design Study}

\vgtcinsertpkg

\title{Inspecting the Process of Bank Credit Rating via Visual Analytics}

\author{Qiangqiang Liu$^{\dag}$\thanks{$\lbrace$codiliu,zhihuazhu,tzye$\rbrace$@tencent.com} %
\and Quan Li\thanks{liquan@shanghaitech.edu.cn; corresponding author}
\and Zhihua Zhu$^{*}$
\and Tangzhi Ye$^{*}$
\and Xiaojuan Ma\thanks{mxj@cse.ust.hk}}
\affiliation{\scriptsize $^{\dag}$ShanghaiTech University$\quad$$^{*}$Corporate Development Group, Tencent $\quad$ $^{\ddag}$The Hong Kong University of Science and Technology}

\teaser{
 \vspace{-4mm}
 \includegraphics[width=\linewidth]{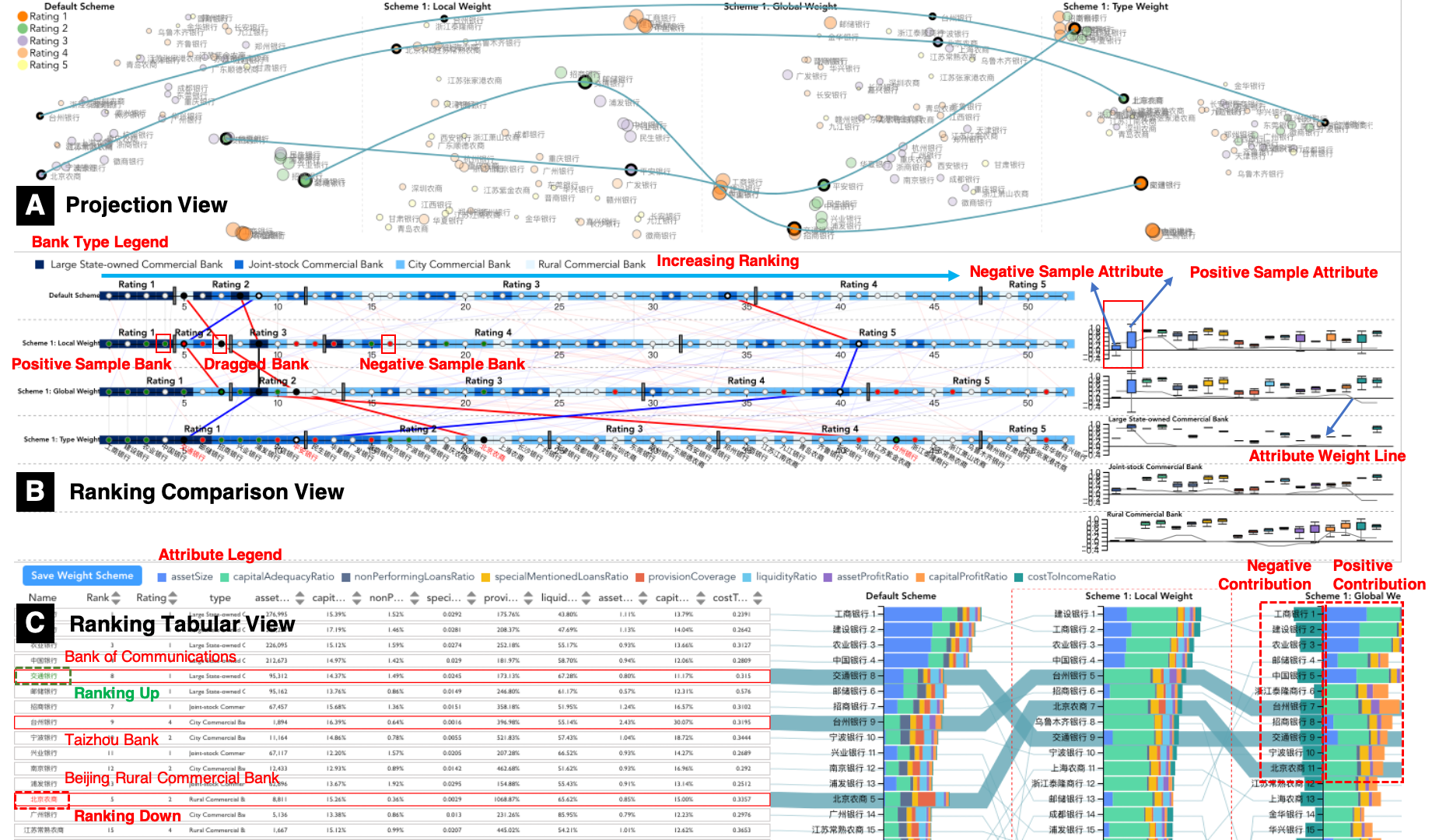}
  \vspace{-6mm}
 \caption{(A) Projection View compares the projections in different ranking schemes. (B) Ranking Comparison View supports analysts to seek explanations of different ranking schemes at an instance level. (C) Ranking Tabular View displays the data attributes, lists the current ranking schemes, and enables analysts to adjust the ranking.}
   \label{fig:teaser}
}

\abstract{
Bank credit rating classifies banks into different levels based on publicly disclosed and internal information, serving as an important input in financial risk management. However, domain experts have a vague idea of exploring and comparing different bank credit rating schemes. A loose connection between subjective and quantitative analysis and difficulties in determining appropriate indicator weights obscure understanding of bank credit ratings. Furthermore, existing models fail to consider bank types by just applying a unified indicator weight set to all banks. We propose \textit{RatingVis} to assist experts in exploring and comparing different bank credit rating schemes. It supports interactively inferring indicator weights for banks by involving domain knowledge and considers bank types in the analysis loop. We conduct a case study with real-world bank data to verify the efficacy of \textit{RatingVis}. Expert feedback suggests that our approach helps them better understand different rating schemes.

} 

\CCScatlist{
  \CCScatTwelve{Human-centered computing}{Visu\-al\-iza\-tion}{}{}
}

\begin{document}


\firstsection{Introduction}

\maketitle

\par Bank credit rating classifies commercial banks into different levels on the basis of the analysis of both the publicly disclosed information and part of the internal bank information according to a certain quantitative standard~\cite{2017Markov,rojas2002rating,treacy2000credit,shen2012asymmetric}. These levels will be succinctly
represented by a combination of numbers and letters, which is convenient for supervisory agencies to conduct an intuitive judgment on the risks of commercial banks and adopt corresponding supervisory measures~\cite{2017Markov,2018Information}. Bank credit rating methodologies have gone through several stages such as subjective judgments, rule-based scorecard analysis, and computational modeling~\cite{gumparthi2011design,brusov2018rating}. Most of these approaches leverage rule-based measures combined with automated methods~\cite{2018Information,Shi2016A} for credit rating assistance. Although these approaches have demonstrated promising performances, domain experts still have the following concerns when directly applying the existing credit rating models to their business scenarios. \textbf{(1) Obscure understanding.} One purpose of the domain experts is to generate an internal bank rating that can be produced consistently by different analysts using the same information with high reliability. Therefore, the methodologies should be intuitive and explainable to benefit from comments and suggestions from the experts advancing the bank credit risk assessment methodologies. \textbf{(2) Loose connection between subjective and quantitative analysis.} Previous bank rating models are extremely biased towards a heavy quantitative approach that requires many quantitative indicators. Therefore, they are not suitable for those imperfect credit systems and banks with limited financial indicators~\cite{2017Markov,2018Information}. Meanwhile, domain experts have their knowledge in the bank rating decision-making process. Integrating domain knowledge and experience with quantitative indicators can improve bank credit rating reliability. \textbf{(3) Unified indicator weights.} Existing credit rating methods apply a unified indicator weight set to all bank entities; however, according to the experts, different types of banks should be assigned to different weights even for the same indicator~\cite{RePEc:bis:biswps:747}. To tackle these issues, we identify domain experts' primary concerns regarding bank credit rating and propose a visual analytics system, namely, \textit{RatingVis} to assist them in exploring and comparing different bank credit rating schemes. Our primary contributions can be summarized as follows: 1) we elicit design requirements of bank credit rating from the literature and an observational study; 2) we propose interactive schemes to infer indicator weights; 3) we design a visual analytics system to help explore and compare different rating schemes.

\section{Related Work}
\par Bank credit rating plays an important role in solving information asymmetry between depositors and banks~\cite{RePEc:pra:mprapa:34864}. For example, CAMEL rating system~\cite{yuksel2015camels} evaluates the \textit{capital adequacy}, \textit{asset quality}, \textit{management}, \textit{earnings}, and \textit{liquidity} of financial institutions, and adopts a five-level scoring system to rate the operation and management level of commercial banks. Moody's bank rating system~\cite{cantor2001moody} focuses on asset quality evaluation by understanding the bank's internal loan approval policies and procedures, internal risk rating methods, the means of managing asset quality and risk to obtain bank-related information and rating data. Standard \& Poor's rating~\cite{peng2002investors} of banks takes into account eight factors such as economic and industry risks, capital, and returns. However, there is no universal global standard for rating. Some methods reply on different data indicators, making it impossible to generalize to other scenarios. Moreover, experts' knowledge of the credit risk is missing in these rating models. Our work integrates domain knowledge into rating and combines quantitative analysis with qualitative evaluation.

\par Entities for rating are inherently multi-attribute data items and several visualization systems that provide interactive multi-attribute data item ranking were proposed~\cite{liu2011learning,cossock2006subset,wall2017podium,zanakis1998multi,10.1111:cgf.13996,puri2020rankbooster}. \textit{ValueCharts}~\cite{carenini2004valuecharts} and \textit{LineUp}~\cite{gratzl2013lineup} allow analysts to create customized rankings by clicking and dragging attributes with adjustable attribute weights. However, they assume that users are able to quantify the importance of particular attributes. To resolve this issue, researchers studied what weight sets lead to a specific ranking. \textit{Podium}~\cite{wall2017podium} allows users to drag the table rows to rank data based on their perception of data value. \textit{WeightLifter}~\cite{pajer2016weightlifter} facilitates the exploration of weight spaces, thereby better understand the sensitivity of a decision to weight changes. The unified attribute weights derived by the above methods are not applicable to our case since banks with various types should be assigned different weights even for the same indicator.

\section{Background and Observational Study}
\par Ensuring the quality of bank rating has attracted attention from many parties. The world-famous rating agencies such as Moody's and Standard \& Poor's evaluate the financial and operational strength and resilience of banks based on operation-related indicators and generate the letter-grade credit. To understand the existing practice of bank rating, we worked with one bank rating expert (E.1, male, age: $32$), one risk management expert (E.2, male, age: $35$), one financial data analyst (E.3, female, age: $24$), and one bank credit expert (E.4, male, age: $28$), who attempted to address the rating problem by ranking. They selected a set of attributes and employed certain multi-criteria decision-making schemes such as \textit{Analytic Hierarchy Process} (AHP)~\cite{al2001application} and \textit{Technique for Order Preference by Similarity to Ideal Solution} (TOPSIS)~\cite{lai1994topsis} in \textit{Statistical Product and Service Solutions} (SPSS) for ranking. Based on the distribution of certain indicators, they divided the ranking into several segments. The banks in a higher up segment are considered better than those in a lower segment. Nevertheless, several issues arise. First, the ranking relies on distance functions to obtain the pairwise similarity. However, different indicators have different numerical scales, thereby requiring normalization. Nevertheless, even the experts who are very familiar with the data have difficulty in estimating the attribute weights after normalization. ``\textit{When we classify banks, we can tell the approximate level of banks by observing certain attributes.}'' However, the subjective perception of attribute importance is ``\textit{difficult to quantify}'', making it challenging to take this intuition to verify the appropriateness of distance functions. Second, banks are of different types. For example, banks in China are classified into ``large state-owned commercial banks'', ``joint-stock commercial banks'', ``city commercial banks'', and ``rural commercial banks''. Generally, the credit rating of the four categories of banks shows a decreasing trend as a whole, but some low-level banks may perform better than high-level banks. In existing bank ratings, experts apply the same weight to the same indicator of different types of banks. However, they realized that ``\textit{the importance of the same indicator among different types of banks should be different.}'' For instance, banks with a high liquidity ratio of assets indicate a strong financing ability, e.g., large state-owned commercial banks have many means to obtain funds while the financing channels of city commercial banks are narrow. Therefore, experts should pay attention to the liquidity ratio of city commercial banks. To sum up, we should meet three requirements: \textbf{R.1 Infer indicator weights by interaction.} Conventionally, the experts rank banks by assigning weights to quantify attribute contributions, by which they cannot efficiently determine which and to what extent certain attributes are important since they only have a holistic understanding of the data. Therefore, they required that the indicator importance could be interactively inferred; \textbf{R.2 Refine indicator weights for different types of banks.} Applying uniform weights to all kinds of banks may result in an irrational credit rating result. As indicated in~\cite{RePEc:bis:biswps:747}, ratings can communicate externally as well the overall riskiness of banks' assets, and ``\textit{different risk weights are assigned to assets with different ratings according to Basel capital rules}''. The experts required an interactive mechanism to refine indicator weights that consider bank types; \textbf{R.3 Facilitate comparison among rating schemes.} The experts wished to preserve previous schemes for comparison so they can understand whether the adjustment leads to a better result. Thus, our approach should facilitate comparison among different schemes.


\section{Back-End Engine}
\par In this section, we derive indicator weights and constraints to generate ranking scores for all items, followed by three ranking schemes.
\par \textbf{Modeling Ranking SVM.} We adopt \textit{Ranking SVM}~\cite{zanakis1998multi,joachims2002optimizing} to derive a set of indicator weights on the basis of the rankings in the data table~\cite{wall2017podium}, which optimizes a SVM hyperplane to the ranking problem with pairwise constraints. Instead of a full set of data with labels, a limited set of pairs of data $d_i$ and $d_j$ and a label are available to derive whether $d_i$ is better or not. The input of the model is difference vectors for data pairs, e.g., $d_i - d_j$, and the output of the model is which point is better~\cite{joachims2002optimizing}. Specifically, \textit{Ranking SVM} transfers the pair of $(d_i, d_j)$ and their relative ranks to a tuple: $d_i - d_j = 1$ if $d_i$ is preferred; otherwise $d_i - d_j = -1$. The generated model predicts which one is better given a pair of points. To avoid an empty result, all constraints are modeled as soft constraints instead of hard ones so that user interaction can always generate a set of attribute weights that model constraints as much as possible~\cite{joachims2002optimizing}.

\par \textbf{Deriving Constraints.} To obtain SVM's linear separator, the ranking problem is transferred into a binary classification problem. Particularly, the labeled data is generated for \textit{Ranking SVM} using the data items with which the user has interacted, e.g., dragging to a new position and there are $k$ marked rows. Without loss of generality, for $k$ points $\{d_{l_1}, ..., d_{l_k}\}$ with indices $[l_1, ..., l_k]$, the set of all combinations of pairwise difference vectors are created as training instances~\cite{joachims2002optimizing}. In other words, for $i, j \in \{1...k\}$, where $i \ne j$, a training tuple is derived, i.e., each training instance is the difference between a pair of rows $d_i$ and $d_j$, classified as  $y=1$ if $d_i$ is ranked higher than $d_j$, or $y = -1$ if $d_i$ is ranked lower than $d_j$. In this study, we set $k = 6$ to ensure a minimum amount of training data for deriving the attribute weight vector after experimental analysis.

\par \textbf{Calculating Ranking Score.} After transforming user interaction and learning the \textit{Ranking SVM} model, a weight vector $\textbf{w}$ is obtained for us to rank the data items. We calculate individual dot products of $w$ with each data item to generate a \textbf{rank score} as:
$
r(d_i) = \textbf{w} \cdot d_i = \sum_{j=1}^mw_jd_{ij}
$ with the highest one corresponding to the top rank.

\par We introduce three ranking schemes based on \textit{Ranking SVM}: \textbf{Scheme 1: Local Weight.} We take original Ranking SVM and only consider the local information as the ranking scheme; \textbf{Scheme 2: Global Weight.} We generate a global weight sequence for all banks. When analysts drag the bank $b_j$ to the position of $k$, we sample a part of banks according to the proportion of each bank type in the original ranking from $1$ to $k-1$ and pair it with $b_j$ to form a positive pair, i.e., the label of $d_q - d_k$ ($q \in [1, k-1]$) is $1$. Similarly, we sample banks to form a negative pair with $b_j$ from $k+1$ to $n$ according to the proportion of each bank type, i.e.,the label of $d_p - d_k$ ($p \in [k+1,n]$) is $-1$. In this way, we generate a general weight set suitable for different types of bank; \textbf{Scheme 3: Type Weight.} When generating pairwise training data, we take bank types into account and generate a set of weights for each bank type. Supposedly we have $b=\{r_1, ..., r_i, ..., r_m\}$ that indicates bank $b$ has $m$ indicators and $s(b_j)$ indicates the ranking score of bank $b_i$ and we define $rankSeq = <b_1, ..., b_j, ..., b_n>$ as an increasing ranking order of banks, i.e., the number of banks is $n$, among which $s(b_1) > ... > s(b_j) > ... s(b_n)$. We separate the ranking of different types of banks in $rankSeq$ to form a ranking for each bank type. When analysts drag a bank $b_j$ to the position of $k$, we check the position of $k$ of $b_j$ in the ranking of each bank type and take $6$ pairs near the position of $k$, e.g., $b_{k-1}$ is better than $b_k$ and $b_{k+1}$ is worse than $b_k$. For each type, we generate a set of weights to calculate the ranking score of the banks of that type.

\par We adapt an entropy discretization method to transfer ranking to rating~\cite{10.1007/978-3-642-40897-7_11,liu2019study}. We first sort ranking scores and treat each score as a segmentation point, and then calculate the entropy of the left and right parts of each point. We consider the division with the minimum entropy as the first division. We repeat the above procedure until we have $5$ divisions (we determine the value of $5$ after discussion with experts). We round down the score of each bank to multiples of $5$. We use $X$ to indicate the random variable of the score and sort the bank scores as $(x_1, x_2, ..., x_n)$. $P(x_i)$ means the probability of score $x_i$. The entropy of $X$ is:
$
H(X)=E[-logP(x_i)]=-\sum_{i=1}^NP(x_i)logP(x_i)
$. Assume there are $k$ different scores among all the ranking scores of banks and $k<n$. We sort the $k$ scores as $(u_1, u_2, ..., u_k)$, which can be considered as breakpoints of the continuous values of $x_1, x_2, ..., x_n$. Then, we choose one point that minimizes the entropy from the candidates. We repeat the process until we have $4$ breakpoints, forming $5$ ratings and complete the discretization.

\section{Front-End Visualization}
\par We propose \textit{RatingVis} to infer indicator weights by interactions and facilitate exploring and comparing bank credit rating schemes.

\par Inspired by \textit{Lineup}~\cite{gratzl2013lineup}, we design \textbf{Ranking Tabular View} to help derive indicator weights from user interaction, evaluate the indicator contribution to the ranking, and support the comparison of different ranking schemes (\textbf{R.1 -- R.2}). As shown in \autoref{fig:teaser}(C), we leverage a table to present the raw data. It displays the name of data items, rank, institution types, and the associated attributes. Analysts can perform a drag-and-drop operation to manually rank data based on their perception of the relative value of the data and domain knowledge. Green indicates the ranking is adjusted higher and red indicates the opposite adjustment; the deeper the color, the more the adjustment. The system then infers a new indicator weight set using the above-mentioned \textit{Ranking SVM} approach. The attained weight vector will be applied to multiply the normalized indicator values for calculating the indicator contribution. The sum of the contributions of all the indicators for one data item, i.e., the rank score, is used to determine the order of the data item. If analysts are satisfied with the current ranking scheme, they can click on the ``Save Weight Scheme'' button and the current ranking scheme will be added on the right side of \autoref{fig:teaser}(C). The associated weight vector will be applied to other views as well. As shown on the right side of \autoref{fig:teaser}(C), we present each ranking scheme as a separate column. These columns adapt bars with different colors to represent different indicators and the length of the bars indicates the contribution of that indicator. To compare different ranking schemes, we arrange them horizontally and connect identical data items across different rankings with blue lines. When selecting a particular data item, a blue bold line connects all the identical data items sequentially. The contributions are divided into positive and negative contributions. Positive contribution (the right side) indicates that the higher the indicator value, the higher the ranking score, and the negative contribution indicates that the higher the indicator value (the left side), the lower the ranking score.

\par \textbf{Projection View} creates low-dimensional projections and preserves local similarities to convey neighborhood structure~\cite{li2018embeddingvis} (\textbf{R.3}). When designing the projection view, we leverage the same weight set and normalized attribute values as used in the ranking tabular view to obtain 2D projection by e.g., \textit{t-SNE}~\cite{2018Multidimensional,sorzano2014survey,van2009dimensionality,maaten2008visualizing} (\autoref{fig:teaser}(A)). Four subviews represent the projections corresponding to the attribute weight vectors from default, local, global, and type weight schemes. The colors indicate different ratings and we encode the bank asset size as the circle size. Analysts can lasso circles on any projection space and all identical ones will be connected via curves.

\par \textbf{Ranking Comparison View.} In previous schemes, different item and the dragged item form pairs of either positive or negative samples for training the Ranking SVM model, allowing analysts to understand what the impacts of different pairs on the ranking result are and how they affect the rating result (\textbf{R.3}). In \autoref{fig:teaser}(B), the ranking differences among schemes are represented by a parallel coordinate-like design, in which each axis represents a ranking scheme and each dot represents a bank. The black dot (dragged bank) is the item dragged by the analyst. Red dots indicate the negative samples while blue ones are the positive samples. We arrange the ranking sequentially (the smaller the better) from left to right and use lines to link all the identical banks across different schemes. For two adjacent ranking schemes, blue lines indicate an increasing ranking, and red lines indicate a decreasing ranking. We encode the bank types as background rectangles with different colors and discretize the ranking into $5$ ratings. On the right side of \autoref{fig:teaser}(B), we use box plots to show indicator distribution of all positive and negative pairs. The color of each box plot represents one indicator. For one indicator, the left box plot indicates the indicator value distribution of negative pairs and the right box plot shows the indicator value distribution of positive pairs. The curve shows the attribute weight.

\section{Case Study}

\par
\begin{figure}[h]
 \vspace{-3mm}
 \centering 
 \includegraphics[width=\linewidth]{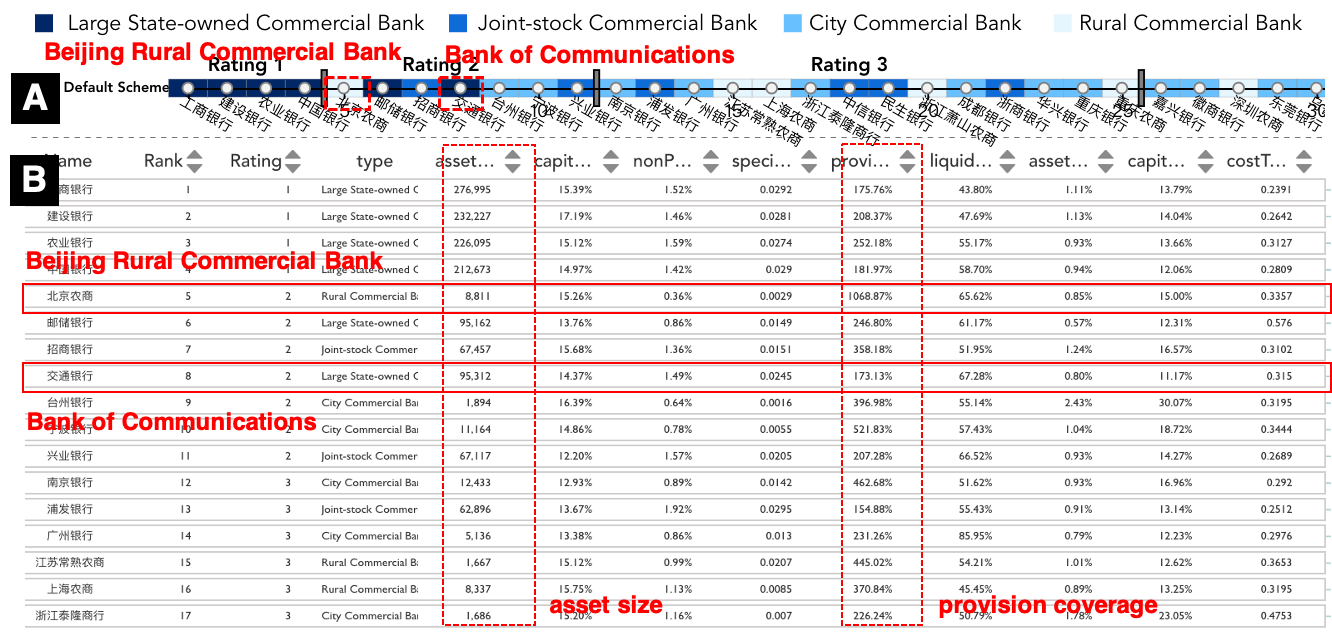}
  \vspace{-6mm}
 \caption{\textit{Beijing Rural Commercial Bank} belongs to ``Rural Commercial Bank'' but ranks higher with rating of $2$. \textit{Bank of Communications} belongs to ``Large State-owned Commercial Bank'' with a low rating.}
 \label{fig:case1}
  \vspace{-3mm}
\end{figure}

 \textbf{Adjusting rankings for some banks.} The experts first observed that \textit{Beijing Rural Commercial Bank} belonging to ``Rural Commercial Bank'' in Default Scheme ranks higher with the rating of $2$ (\autoref{fig:case1}(A)). \textit{Bank of Communications} belongs to ``Large State-owned Commercial Bank'' but has a lower rating of $2$. From \autoref{fig:case1}(B), the experts witnessed that although the \textit{asset size} of \textit{Beijing Rural Commercial Bank} is low ($8,811$), the \textit{provision coverage} is high ($1068.87\%$), and that is why it is relatively high in the ranking. On the contrary, although the \textit{asset size} of \textit{Bank of Communications} is relatively high ($95,312$), its \textit{provision coverage} is much lower ($173.13\%$), leading to a low ranking. The experts commented that according to the bank's previous performance and their expertise, the current ranking of \textit{Beijing Rural Commercial Bank} is higher than expected, although it has excellent \textit{provision coverage}, the \textit{asset size} is worse than that of the banks ranked near it. Similarly, the ranking of \textit{Bank of Communications} is also lower than expected. ``\textit{After all, Bank of Communications is one of the six most famous state-owned banks in China,}'' said E.1, and ``\textit{its asset size is strong}''. The experts then manually adjusted the ranking of the two banks, ``\textit{I will drag the bank to a ranking that meets my psychological expectation and also consider the bank type.}'' E.1 witnessed that the banks ranked $15$ and $16$ are ``Rural Commercial Bank'' which belong to the same bank type with \textit{Beijing Rural Commercial Bank}. Meanwhile, E.1 thought that \textit{Beijing Rural Commercial Bank} is better than them (banks at the position of $15$ and $16$) and is also better than \textit{Guangzhou Bank} (at the position of $14$) in terms of \textit{asset size}. Therefore, E.1 decided to move \textit{Beijing Rural Commercial Bank} from its current position of $5$ to $13$ and \textit{Bank of Communications} from its current position of $8$ to $5$. They then clicked on the ``Save Weight Scheme'' button and the system generated a new weight set, which is applied to generate three new rankings. The adjusted result is shown in \autoref{fig:teaser}(C).

\par
\begin{figure}[h]
 \vspace{-3mm}
 \centering 
 \includegraphics[width=\linewidth]{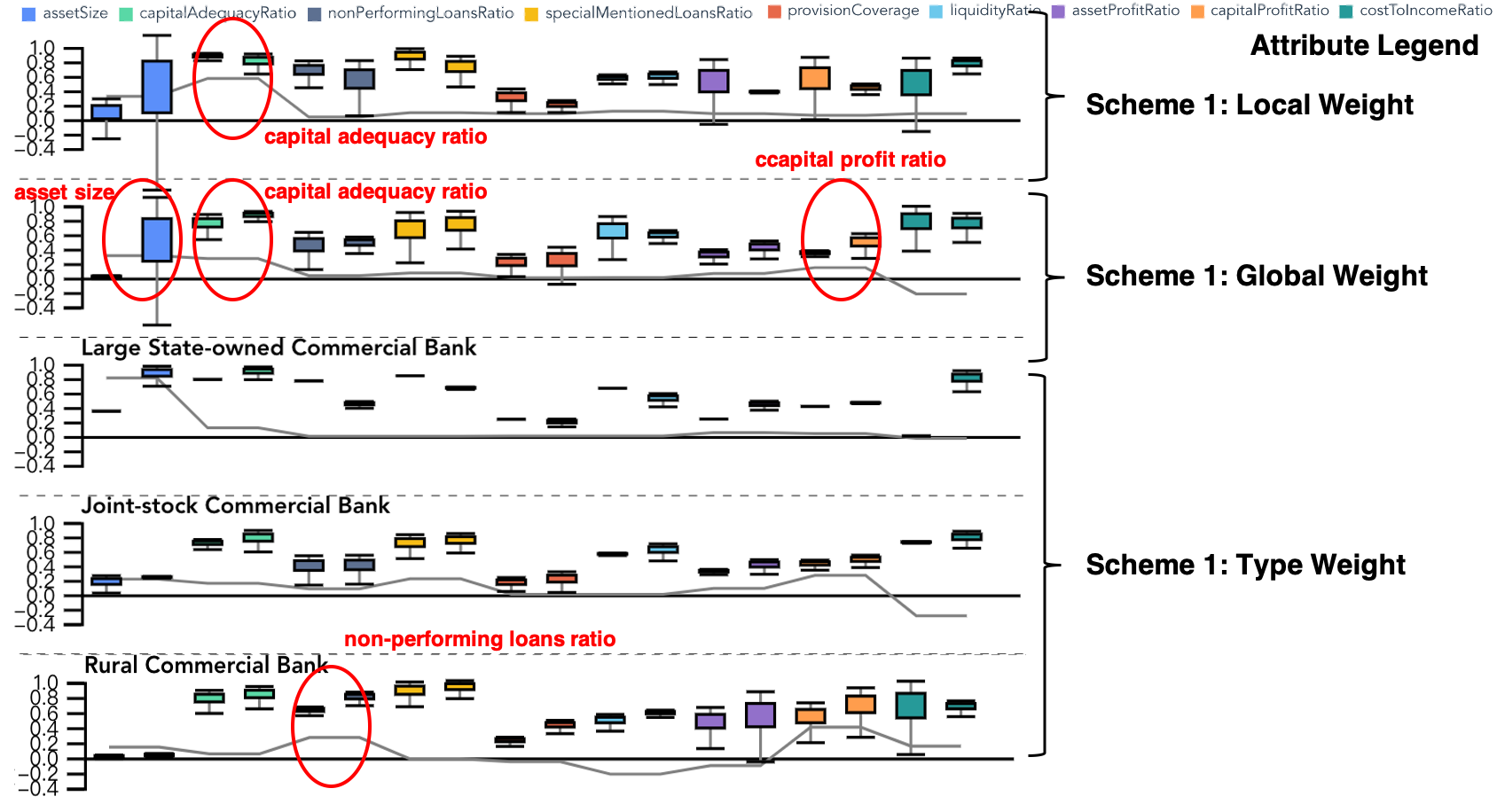}
  \vspace{-6mm}
 \caption{\textit{Capital adequacy ratio} is more important in Local Weight. In Global Weight, \textit{asset size}, \textit{capital adequacy ratio}, and \textit{capital profit ratio} are emphasized. In Type Weight, \textit{asset size} and \textit{non-performing loans ratio} is mainly considered in large state-owned commercial banks and rural commercial banks, respectively.}
 \label{fig:case3}
   \vspace{-3mm}
\end{figure}

 \textbf{Comparing ranking schemes.} The experts found that in Type Weight some banks with a rating of $2$ are close to the banks with a rating of $1$ but they are not very close in the ranking tabular view (\autoref{fig:teaser}(A)). With curiosity, E.1 selected \textit{Beijing Rural Commercial Bank} in \autoref{fig:teaser}(C), and observed the ranking differences in different ranking schemes and how indicator contribution affects the overall ranking in each scheme. In \autoref{fig:case3}, the experts further identified that due to different schemes of sampling positive and negative pairs, the value distribution (the distance between the upper and lower edges of the box plot) in Local Weight and the height of the box plot are more concentrated than those in Global Weight, which makes sense since in Local Weight, both positive and negative samples are selected near the dragged bank, compared with the random selection of positive and negative samples in Global Weight. Particularly, in Local Weight, indicators such as \textit{capital adequacy ratio} are more important while in Global Weight, indicators such as \textit{asset size}, \textit{capital adequacy ratio}, and \textit{capital profit ratio} are emphasized.

\par
\begin{figure}[h]
 \vspace{-3mm}
 \centering 
 \includegraphics[width=\linewidth]{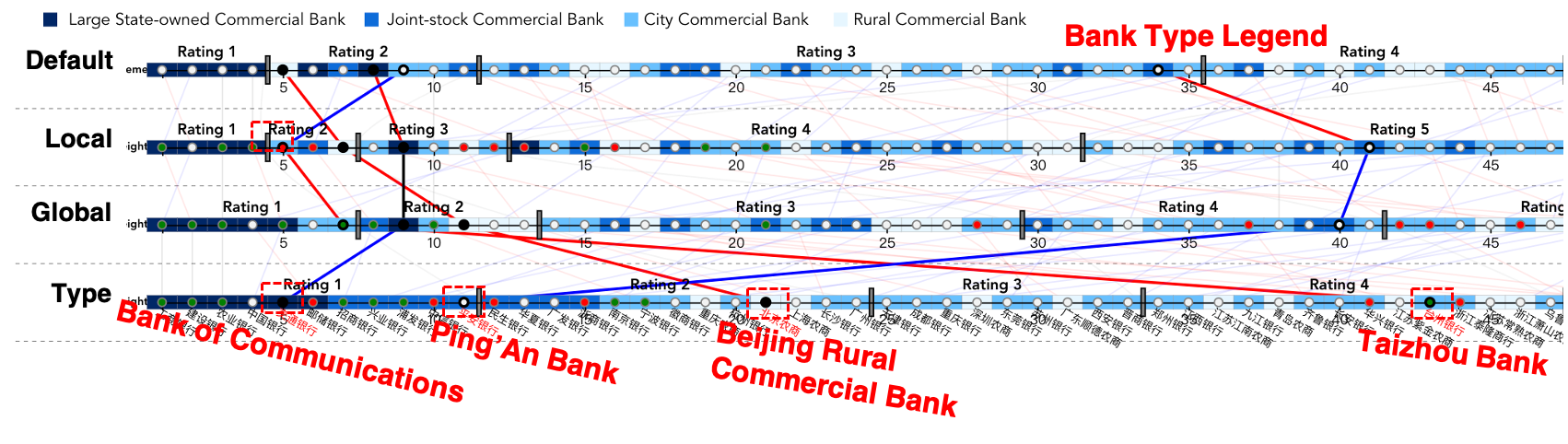}
  \vspace{-8mm}
 \caption{Ranking comparison of four banks among Default Scheme, Local/Global/Type Weight rankings.}
   \vspace{-3mm}
 \label{fig:case2}
\end{figure}

 \textbf{Identifying critical indicators for ranking different types of banks.} From \autoref{fig:case2}, in Default Scheme, \textit{Beijing Rural Commercial Bank} belongs to \textit{Rural Commercial Bank} and it is located around large state-owned commercial banks, which is not reasonable. From \autoref{fig:case3}, the experts confirmed that the indicator weights of large state-owned commercial banks mainly lie in \textit{asset size}, while for rural commercial banks, the \textit{non-performing loans ratio} matters. E.1 explained ``\textit{smaller banks tend to lend more loans with risks, leading to a high non-performing loans ratio, while the risk management of large banks is good and lending is conservative. So non-performing loans ratio is smaller}''. Thus, the \textit{non-performing loans ratio} are not significant for distinguishing large state-owned commercial banks, leading a smaller weight. The experts further noticed that in the axis of Type Weight, \textit{Bank of Communications} has a rating of $1$, ranking the $5^{th}$ and \textit{Beijing Rural Commercial Bank} has a rating of $2$ with ranking $21^{st}$; however, in Default Scheme, \textit{Bank of Communications} has a rating of $2$, ranking the $8^{th}$ and \textit{Beijing Rural Commercial Bank} has a rating of $2$, ranking $5^{th}$. This observation demonstrates that the performance of \textit{Bank of Communications} in Type Weight is improved while the performance of \textit{Beijing Rural Commercial Bank} is worse, which is consistent with the adjustment by the experts, i.e., E.1 adjusted the ranking of \textit{Beijing Rural Commercial Bank} to the $13^{rd}$, although in Type Weight, the ranking is recommended as the $21^{st}$. The experts also observed that the ranking of \textit{Taizhou Bank} varies among different schemes. As shown in \autoref{fig:teaser}(C), E.1 found that although the performance of \textit{Taizhou Bank} in \textit{asset size} is on average, it performs well in terms of \textit{provision coverage}, \textit{asset profit ratio}, and \textit{capital profit ratio}, leading to overall good performance in Default Scheme (\autoref{fig:case2}). Similarly, the rankings of \textit{Taizhou Bank} in Local Weight and Global Weight are all highly ranked. However, E.1 thought the rankings of \textit{Taizhou Bank} were seriously overestimated in the first three schemes and he was satisfied with the ranking in Type Weight. On the contrary, the experts identified the rankings of \textit{Ping'An Bank} in the first three schemes were underrated (\autoref{fig:case2}), but ``\textit{it actually has a good credit rating,}'' said E.1. He commented that the rating in Type Weight (rating of $2$) is more in line with his expectation, confirming his intuition that ``\textit{the weights of different bank types should be treated separately.}''

\section{Conclusion and Future Work}
\par We propose \textit{RatingVis} to facilitate domain experts exploring and comparing different bank credit rating schemes. A ranking tabular view deduces the indicator weights based on user interaction; a projection view visualizes the ranking data distributions generated by different ranking schemes, and a ranking comparison view compares the ranking schemes at an instance level. A case study verifies the efficacy of \textit{RatingVis}. In the future, we will consider attributes with various types and explore more ranking schemes for comparison.
\balance

\acknowledgments{
We thank the anonymous reviewers. This work is partially supported by the research start-up fund of ShanghaiTech and HKUST-WeBank Joint Laboratory Project Grant No.: WEB19EG01-d.}

\bibliographystyle{abbrv-doi}

\bibliography{template}
\end{document}